# Dynamic Construction of Belief Networks

Robert P. Goldman and Eugene Charniak*

Dept. of Computer Science, Brown University
Box 1910,
Providence, RI 02912

## Abstract
We describe a method for incrementally constructing belief networks. We have developed a network-construction language similar to a forward-chaining language using data dependencies, but with additional features for specifying distributions. Using this language, we can define parameterized classes of probabilistic models. These parameterized models make it possible to apply probabilistic reasoning to problems for which it is impractical to have a single large, static model.

## Introduction
We describe a method for incrementally constructing belief networks. We have developed a network-construction language similar to a forward-chaining language using data dependencies, but with additional features for specifying distributions. Using this language, we can define parameterized classes of probabilistic models. These parameterized models make it possible to apply probabilistic reasoning to problems for which it is impractical to have a single large, static model.

There are a number of applications where such a facility would be useful. An expert system for genetic counseling would be able to quickly draw up and evaluate a pedigree network for a given family. Our language provides a convenient facility for combining general knowledge from a domain theory (in this case, genetics) with facts about a specific case (an individual pedigree). One could also diagnose machine faults in a large class of machines, all built out of a given set of components. For example, one could define a class of models for diagnosing mechanical problems in motorcycles. This class of models would defined in terms of the operation of components. To create a instance of this class of models for a particular motorcycle, one would specify the values of parameters like fuel supply (carburetion or fuel-ignition), transmission (chain or shaft), number of cylinders, etc. For pattern theory, this approach would make it more convenient to experiment with different topologies and bond functions.[1]

This technique is tailored to the production of belief networks that can be incrementally constructed and evaluated. We have been working with belief networks which are theoretically infinite. For computational practicality it is necessary that we be able to confine our attention to small, directly relevant portions of such networks. For such situations, our language provides the ability to compose different probabilistic influences in stereotyped ways.

In the next section of this paper, we will give a (very) brief introduction to belief networks, the graphical representation we use for probability distributions. Then we will give an outline of our language for belief network construction. We will illustrate the use of our language with a simple example from the domain of genetics. Then we will outline the application of dynamically-constructed belief networks to the problem of natural-language understanding, and the way it has directed the evolution of our network-construction language. Finally, we will discuss related work.

## Belief Networks
Belief networks are directed acyclic graphs used to represent probability distributions.[2] In a belief network, nodes represent random variables, and arcs represent direct influences between random variables. The probability distribution corresponding to a belief network may be specified by giving a conditional probability table for each node, conditioned on its parents. This local representation makes possible the rule-based specification of probability distributions that we describe here.

There are other graphical representations for probability problems that are also amenable to our approach. Belief networks are the most suitable to the natural-language problems we face. For problems in

*This work has been supported in part by the National Science Foundation under grant IRI-8911122, and Office of Naval Research under grant N00014-88-K-0589.

---

[1] An observation for which we are indebted to Ulf Grenander.

[2] Judea Pearl's book [Pearl, 1988] gives a thorough account of the properties of such networks.

decision-theory, influence diagrams would be appropriate [Schachter, 1986]. For problems in pattern theory, Markov networks may be used. Because we have some nodes between which it is difficult to assign a causal direction, we are considering using chain graphs, which mix directed and undirected edges [Lauritzen, 1988, Frydenberg, 1989]. Our techniques would be equally applicable to these other representations; it would just be necessary to adapt the way the local probability functions are specified.

## Our network-construction language

This section describes how rules are written in our network-construction language, FRAIL3. FRAIL3 has the following features:

- It allows us to create propositions, random variables in a belief network, which are also indexed as statements in a first-order logic database.

- In order that these belief networks be solved, FRAIL3 makes it convenient to specify the conditional probability matrices of these propositions, using declarative knowledge stored in the logical database.

Our probabilistic database has a great deal in common with a deductive database with truth maintenance (see [Charniak and McDermott, 1985] for an introduction, or [Doyle, 1979, McAllester, 1982, deKleer, 1986]). Such a system combines a logic-programming language with forward- and backward-chaining (bottom-up and goal-directed inference, respectively) with a data dependency, or justification, network. When statements are added to the database, they are given a justification, which records the antecedent statements necessary to justify belief in the consequent.

Our database language is similar to a TMS language in having rules for forward and backward reasoning. It differs in allowing greater flexibility in specifying the way arcs connecting database statements are added. In a TMS, one always adds arcs (justifications) from antecedent statements to consequents. However, in evidential reasoning, we often want to chain from observed statements to possible causes, and direct arcs from the causes to the observation. Our rules allow us to specify independently the statements to be added to the database, and the arcs to be drawn between them.

The data structures corresponding to the arcs are also more expressive. Rather than having justifications, we attach *pforms* to the statements in the database. Pforms contain information which specifies the probabilistic dependency of the node at the head of an arc set on its direct ancestors. More formally, each pform is a hyperedge from a set of statements to one statement. This hyperedge corresponds to a set of edges in a belief-network. See Figure 1

The important operators of our database language are →, ←, →← defpreddist and index In the next several

:parents (a, b, c) :child v
:parents (d, e) :child v

(a) pforms

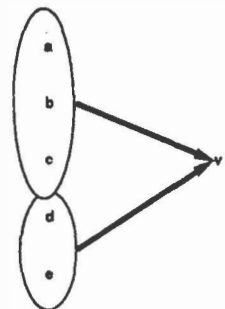

(b) corresponding hypergraph

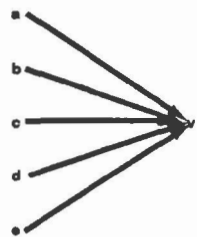

(c) corresponding belief network

Figure 1: The interpretation of pforms.

paragraphs we will discuss each of these operators and how they work.

(← *query antecedent(s)*) To answer a request for *query*, retrieve *antecedent(s)*. A conventional backward-chaining rule.

(→ *trigger consequent* :prob *pforms*) → is the basic network-constructing operator. When a statement unifying with trigger is added to the database, apply the bindings to consequent, and add it to the database. The pforms control the way arcs are added to the network, and give information about the way the conditional probability matrices are set up at the tails of the arcs. This information will be used by the distribution functions discussed below. A sample rule:

**Rule 1**
(→(condition ?x) :label ?caused
  (causal-event ?x) :label ?cause
  :prob ((?cause →?caused ((t |t) .9) ((t |f) :p))))

This states that whenever we predicate condition of some entity, we should also postulate a corresponding causal-event. The first part of the :prob argument specifies that we should draw an arc from the causal-event node to the condition node. In order that we be able to specify the addition of arbitrary arcs, we need to be able to refer to statements by name. The :label operator provides this capability. In this rule, ?caused will be bound to a pointer to the triggering statement, and

?cause will be bound to a pointer to the newly-added statement. It should be emphasized that, unlike in a conventional TMS language, which automatically adds arcs from antecedents to consequents, we do *not* add any arcs other than those explicitly designated by :prob arguments to rules.

The second part of the :prob argument indicates that when we draw up the conditional probability matrix for the condition node, its probability given that causal-event is true is .9. Its probability given that causal-event is false should default to the prior probability (the way this defaulting can be done will be discussed below when we discuss defpreddist).

(→←  *additional-antecedents consequent pforms*) If FRAIL3 is told to add a statement of this form to its database, it will try to retrieve *additional-antecedents*, and apply their variable bindings to assert *consequent*. This operator is often used to retrieve additional facts from our domain knowledge. For example, the following is a rule used in our language-understanding application, for reasoning about word-senses:

Rule 2
(→(word-inst ?i ?word) :label ?A
    (→←(word-sense ?word ?frame ?prob)
        (inst ?i ?frame) :label ?C)
  :prob ((?C →?A) ((t | t) ?prob) ((t | f) (/ :p 100))))

This rule may be read as follows: If a node is added to the network describing a new word-token (an instance of a word), and if there is a statement in the database specifying that one of the senses of this word is ?frame, add a node for an instance of the type ?frame. Draw an arc from this newly-added node (?C), to the word-inst node (?A). When drawing up the probability matrix for the word-inst node, use information about the frequency with which it is used to express the concept frame, for the case where the node for frame is true, else use a default value of $\frac{1}{100}$th of the prior probability.

Another common use of →←  is to collect additional parents for a node. For example, one of the tasks for our natural language system is to find referents for pronouns. A desire to refer to an already-discussed person or thing can be the explanation for using a pronoun. So when we see a pronoun, we need to find all the previously known entities that are of the same gender as the pronoun, and build arcs from them to the newly-added statement. This requires the kind of combined forward- and backward-chaining that →←provides. A simplified version of our pronoun rule follows:

Rule 3
(→(word-inst ?i ?pronoun) :label ?PRONOUN
    (→←(and (inst ?x ?frame) :label ?POS-REFERENT
            (compatible-gender ?pronoun ?frame))
        (has-referent ?i) :label ?HAS-REF)
  :prob (?POS-REFERENT →?HAS-REF
            ((t |t) (ref-prob ?frame ?pronoun)
            ((t |f) 0)))

(?HAS-REF →?PRONOUN)
    (((t |t) 1) ((t |f) .00001)))

(**defpreddist** *predicate prior-fn posterior-fn*) These rules are declarations, which tell us what functions to use to set up the conditional probability matrices for statements with predicate *predicate*. If such a statement is a root node of the graph, *prior-fn* will be called with the statement as an argument. Otherwise, FRAIL3 will apply *posterior-fn* to the pforms attached to the statement. The function of the posterior-fn is to combine the information from the various pforms.

We need **defpreddist** because our other rules are not sufficient to quantify belief networks. There are two cases where we fall short:

1. They do not allow us to specify distributions for nodes without parents, and

2. they do not dictate how to fill out the conditional probability distributions for nodes which have more than one pform.

Problem (2), filling out the conditional probability distributions, arises in hand-construction of belief networks, as well. It occurs because, typically, rather than being able to determine the probability of a random variable conditioned on all direct influences, one only has lower-order conditional probabilities.

Functions which we use as posterior-fns in our language-processing application are noisy versions of various logic gates: noisy-or, noisy-xor and noisy-and. To see how these are used, consider the rule for word-senses, given above as rule 2. The posterior-fn we use for word-inst statements is xor-dist. Typically, words are ambiguous, and have a number of senses. Each such sense will give rise to a different pform. What we want our posterior-fn to dictate is that each one of these senses is a possible explanation for the use of the given word, and that the senses are mutually exclusive. Pseudo-code for xor-dist is given as Figure 2. This is a simple procedure which specifies that the probability of a node given a state of its parents is 1 if that conditioning-case satisfies one and only one pform, and 0 otherwise.[3]

(**index** *statement*) Adds statements to the first-order logic database that are *not* also linked into the belief network. Rules, and other non-ground formulae are indexed.

## An example from genetics

To give an idea of how our system can be used, we give our treatment of an example in genetics. We have adapted this example from [Jensen *et al.*, 1988] and [Spiegelhalter, 1988]. This example has been chosen

---

[3] A conditioning-case is a list of (node . state) pairs, and a node-case is (node . state). This terminology is borrowed from IDEAL[Srinivas and Breese, 1989].

```
procedure xor-dist (node pforms)
/*this loops through all possible instantiations of the predecessors
  of node */
for conditioning-case = all-states(node-predecessors node)
   /* this loops through all possible assignments to node --
      in this case, since node is boolean, :true and :false */
   for node-case = all-states(node)
      number-true = number of pforms satisfied by conditioning-case;
      if state-name(node-case) = :true then
         if number-true = 1 then
             prob[node-state,conditioning-case] = 1;
         else
             prob[node-state,conditioning-case] = 0;
      else %the state of the node is :false
         if number-true = 1 then
             prob[node-state,conditioning-case] = 0;
         else
             prob[node-state,conditioning-case] = 1;
   end for node-case;
end for conditioning-case;
end procedure.
```

Figure 2: Pseudo-code for the xor-dist posterior-fn.

solely for its convenience and clarity. Any errors should be attributed to our ignorance about the domain.

We consider a gene with two allelles (or values), a1 and a2. Every individual's **genotype** consists of two alleles, one inherited randomly from each parent. Because we have no way, in general, of knowing from which parent a given allele has been inherited, we represent genotypes as unordered pairs: a1a1, a1a2 or a2a2. a1 is recessive and gives rise to a detectable condition, a2 is dominant and does not. Therefore, the **phenotypes** (the detectable conditions) are **present** corresponding to genotype a1a1, and **absent** corresponding to genotypes a1a2 and a2a2.

In order to model the situation resulting in populations with these genotypes, we need to know probabilities determining the transmission of the genes from parents to children. These are given in Figure 3. In order to avoid infinite regress, we must make some assumptions about the genotypes of individuals about whose ancestry we are ignorant (**founders**). We assume they are members of a population in Hardy-Weinberg equilibrium. This yields the follwing table of priors:

$$\begin{aligned} p &= p(a1) \\ q &= p(a2) = 1-p \\ p(a1a1) &= p^2 \\ p(a1a2) &= 2pq \\ p(a2a2) &= q^2 \end{aligned}$$

Finally, we need the **penetrance probabilities**: the distribution of the phenotype given a genotype. A very simple one is given in Figure 4.

The rules necessary to describe this situation are:

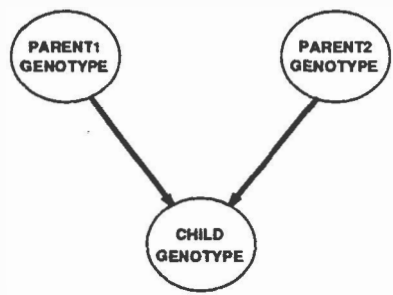

|      | a1a1          | a1a2              | a2a2          |
|------|---------------|-------------------|---------------|
| a1a1 | (1, 0, 0)     | (0.5, 0.5, 0)     | (0, 1, 0)     |
| a1a2 | (0.5, 0.5, 0) | (0.25, 0.5, 0.25) | (0, 0.5, 0.5) |
| a2a2 | (0, 1, 0)     | (0, 0.5, 0.5)     | (0, 0, 1)     |

Figure 3: Conditional Probability Matrix for Transmission Probabilities. The three entries in each cell are P(a1a1|*conditioning-case*), P(a1a2|*conditioning-case*), and P(a2a2|*conditioning-case*), respectively. Genotypes of parents are the horizontal and vertical indices.

**Rule 4** *Transmission of genotypes:*

(→(child ?c ?p1 ?p2)
    (and (genotype ?c) :label ?CHILD-NODE
         (genotype ?p1) :label ?PARENT-1
         (genotype ?p2) :label ?PARENT-2)
    :prob ((?PARENT1 ?PARENT-2 →?CHILD-NODE
        ((a1a1 |a1a1 a1a1) 1)
        ((a1a1 |a1a1 a1a2) 0.5) ((a1a2 |a1a1 a1a2) 0.5)
        ((a1a2 |a1a1 a2a2) 1)
        ((a1a1 |a1a2 a1a1) 0.5) ((a1a2 |a1a2 a1a1) 0.5)
        ((a1a1 |a1a2 a1a2) 0.25) ((a1a2 |a1a2 a1a2) 0.5)
        ((a1a2 |a1a2 a2a2) 0.5) ((a2a2 |a1a2 a2a2) 0.5)
        ((a1a2 |a2a2 a1a1) 1)
        ((a1a2 |a2a2 a1a2) 0.5) ((a2a2 |a2a2 a1a2) 0.5)
        ((a2a2 |a2a2 a2a2) 1))))

**Rule 5** *Penetrance:*

(→(observed-phenotype ?NAME ?CONDITION)
    (→←(penetrance-prob ?PROB)
        (and (genotype ?NAME) :label ?GENOTYPE
             (phenotype ?NAME) :label ?PHENOTYPE
             (instantiate-observation ?PHENOTYPE
                ?CONDITION)))
    :prob ((?GENOTYPE →?PHENOTYPE
        (((present |a1a1) ?PROB)
         ((absent |a1a2) 1) ((absent |a2a2) 1)))))

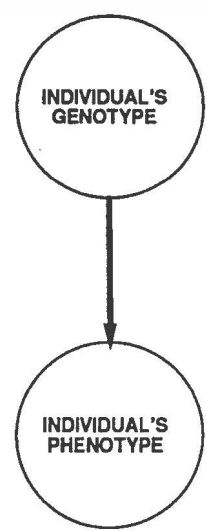

Figure 4: Conditional Probability Matrix for Penetrance Probabilities.

|  | Genotype | | |
|---|---|---|---|
| Genotype | a1a1 | a1a2 | a2a2 |
| :present | 1 | 0 | 0 |
| :absent | 0 | 1 | 1 |

The instantiate-observation statement above will not be a real statement. We specify an assert function for this predicate, so that when FRAIL3 is told to assert such a statement, it will instead call the function add-evidence on the statement to which ?PHENOTYPE is bound.

**Rule 6** *Assert function for* instantiate-observation:

(defun assert-observation (pattern)
    (add-evidence (second pattern)
        (third pattern)))

(setf (assert-function instantiate-observation)
    assert-observation)

**Rule 7** *Penetrance probability:*
(penetrance-prob 1)

**Rule 8** *Predicate definitions:*

(defpreddist genotype hardy-weinberg transmission)
(defpreddist phenotype nil simple-pform)

We should note a couple of features of the above rules. First of all, note that we can use the same rules for many different values of $p$ and $q$, simply by changing the prior-function for genotype nodes. Furthermore, we can change the penetrance probability just by changing rule 7. Finally, note that this scheme avoids adding unnecessary phenotype nodes. We will only have phenotype nodes which correspond to actual observations.[4]

In [Jensen *et al.*, 1988], Jensen et. al. give an example of a belief network for an example of incest. It is given as Figure 5. Individual E is known to have the condition. We would create a diagram corresponding to this situation by asserting the following statements into a database containing the above rules:

(child C A B)
(child D B F)
(child E C D)
(observed-phenotype E :present)

It can be seen from the above how easy it is to use FRAIL3 to combine information about a class of situations with the details of a particular case to construct a complete probabilistic model.

### How we use dynamic belief networks

Our interest in belief networks has grown out of our work on story understanding. The problem of story understanding is to take a story written in some natural language, and "understand" it in the sense that one's program could answer questions about it. It is generally understood that this involves translating the story into an internal representation which will support the inferences necessary to answer questions. This internal representation should be an unambiguous representation of the meaning of the text. In this translation process, the program will make use of knowledge about language and about the world the language is being used to describe. Central to the problem of story understanding is plan-recognition: the ability to recognize the plans and goals of characters in a story, based on their actions.

---

[4] If we so desired, we could also have phenotype nodes for individuals about whom we would like to query.

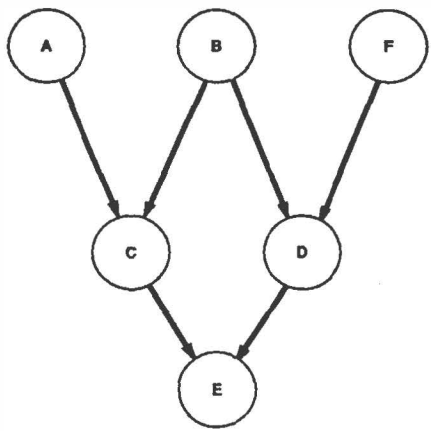

Figure 5: Belief network/pedigree corresponding to a case of incest.

For some time we have been interested in the problems posed by uncertainty in story understanding. Uncertainty in language understanding comes from the ambiguity which exists at all levels of natural-language: syntactic, semantic and pragmatic. Our current program, Wimp3, operates by turning problems of text interpretation into probability questions such as "What is the probability that this word has this sense, given the evidence?" (where the evidence is the text of the story so far), and "What is the probability that the action described by the word plays a role in the actor's plan to achieve some known goal?" Wimp3 does this on a word-by-word basis. These probability questions are formulated as belief networks.

The networks we create contain nodes describing the text at all many levels of detail (see Figure 6 for an example). At the leaves are the nodes describing the 'evidence' we've observed: the input string. Above these are nodes describing the denotations of the words in the input (e.g., **liquor-store3** in Figure 6). At the top of the networks are descriptions of the plans of the characters in the story (*liquor-shop4* is a hypothesized plan to buy liquor at a liquor-store).

More precisely, Wimp3 works as follows:

1. A parser reads a single word of the English text. It produces statements which describe the words in the story and the syntactic relations between them.
2. The output of the parser is used by the network construction component. This component contains rules for language abduction. It builds a net, or extends the current net if some input has already been received.
3. We collect the portion of the belief network which may be affected by the most-recent additions to the belief network. This sub-graph is evaluated by a network-evaluation component.[5]
4. If certain conclusions are overwhelmingly favored, they may be accepted as true to simplify further computation. Statements not rejected as too improbable will trigger further network-expansion. If nodes are added to the network, go to step 3, else, return to step 1.

To return to the example of Figure 6, on the first iteration after reading 'liquor-store' we will add the statement (inst liquor-store3 liquor-store) to the database. This hypothesis will not be eliminated as false when the network is evaluated, so we will consider it further, adding the nodes for a liquor-shopping plan. We do this iterative expansion of the network to keep its size small. For example, one possible (although unusual) use of the word 'went' is to mean 'died', as in "Franco went at 10 o'clock last night." In the context of "Jack went to the liquor-store." this sense is clearly ruled out by syntactic information. By expanding the belief network incrementally, we avoid adding nodes to represent explanations for the non-existent event of Jack dying.

We use the IDEAL system [Srinivas and Breese, 1989] to evaluate the networks we Wimp3 creates. IDEAL is a program allowing the specification of belief networks and influence diagrams in LISP. It provides a framework within which a number of network-solution algorithms can be applied to such diagrams. Wimp3 creates net-works in FRAIL3, then decides (see step 3 above) what portion of the belief network has changed and needs to be evaluated. This part of the network is solved by a version of Jensen's junction-tree algorithm [Jensen, 1989], which we have optimized for special features of our networks[Goldman and Boddy, 1990].

---

[5]This aspect of the work has not been thoroughly worked-out. At the moment, we simply commit to beliefs based on the probability a statement has after a fixed period of time.

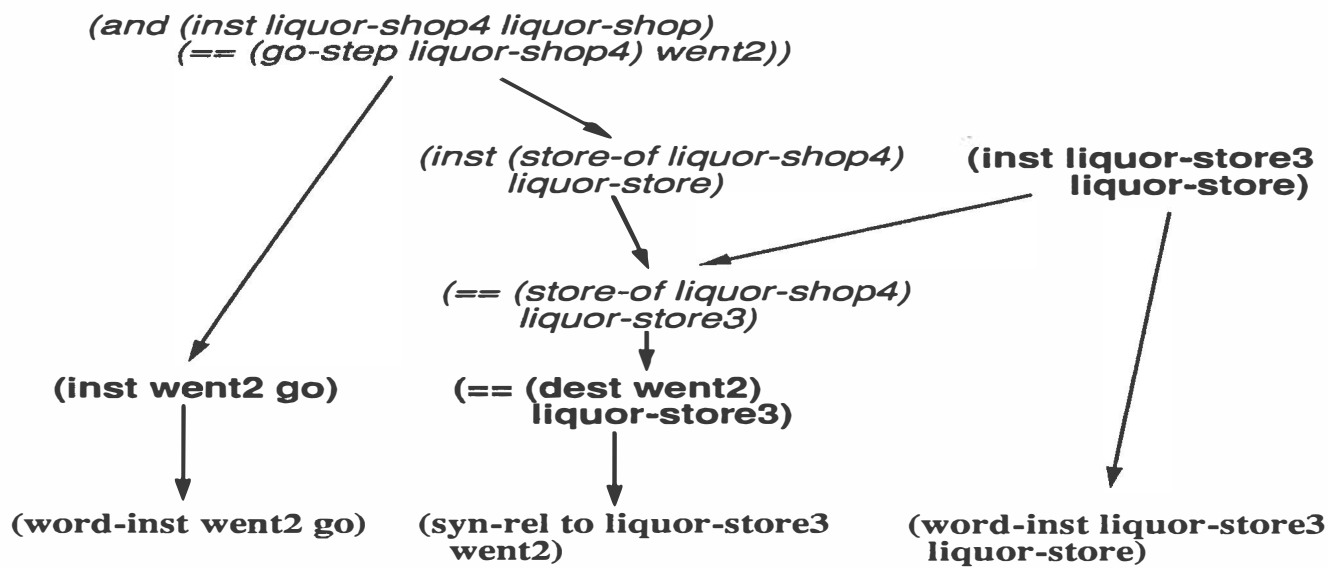

Figure 6: A fragment of the belief network representation of the sentence "Jack went to the liquor-store."

## Related Work

There has been a great deal of work on probabilistic ATMS's for dynamically constructing probabilistic models. An ATMS[deKleer, 1986] is a multiple-context truth maintenance system. Laskey and Lehner [Laskey and Lehner, 1988] have shown that Dempster-Shafer theory can be mapped onto the ATMS in a straightforward way. D'Ambrosio developed a probabilistic ATMS[D'Ambrosio, 1988a, D'Ambrosio, 1988b], and we developed a probabilistic ATMS tailored for language understanding[Goldman and Charniak, 1988].

In our experience, these logical-probabilistic hybrids are unsatisfactory for a number of reasons. Probabilistic inference and logical inference do not correspond perfectly. The operation of conditioning is not natural in logic. Logical inference does not permit us to distinguish between evidential and causal support for a proposition. Pragmatically, it seemed wasteful to combine two expensive labelling procedures – ATMS context labelling and probabilities.

We are greatly indebted to Judea Pearl for his idea of combining causal influences in stereotyped ways like the noisy-or gate [Pearl, 1988]. Our posterior-functions are a generalization of this idea.

In work done simultaneously with ours, Jack Breese has developed very similar techniques for rule-based construction of probabilistic models[Breese, 1989]. The differences between his work and ours arise from the different problems we are trying to solve. Breese's method is aimed at creating belief networks and influence diagrams in order to reply to queries. Because our work is aimed at interpretation, and requires complex hypotheses, we do not have clearly goal-directed networks.

Another difference is that Breese does not have anything corresponding to our posterior-functions. Associated with his network-constructing rules are fully-fleshed-out conditional probability distributions. When more than one such distribution applies, his system chooses which to apply according to some simple heuristics. We need to be able to specify in greater detail how to compare different probabilistic dependencies, because our models are too big to describe more explicitly, and because nodes and arcs are added to and deleted from our networks.

Two examples will clarify this important difference: First of all, consider our rule for finding word-senses (rule 2). If each word has a large number of possible senses, it is clearly impossible for us to store a full conditional probability matrix for each word, given all its senses. Impossible, and unnecessary as well, because the different causes interact in a stereotyped way. Second, our interpretation problem is hierarchical: we make hypotheses about actions and events to explain the text, and then consider the plans and goals of characters in the story in order to explain these actions and events. We do not want to expand our networks to include plans hypothesized to explain characters' actions until we are at least reasonably sure that the actions we are trying to explain actually occurred – that we have not simply misunderstood some verb. This leads to our need to make root nodes into interior nodes.

More details on our work on language understanding is given in [Charniak and Goldman, 1989a, Carroll and Charniak, 1989, Charniak and Goldman, 1989b, Goldman and Charniak, 1990].

## Conclusions

In this paper we have outlined a language for constructing belief networks on an as-needed basis. This language of rules makes it convenient to define parameterized classes of probabilistic models. Such parameterized models are more convenient to use, and more efficient to represent than full-blown belief networks. We have demonstrated the usefulness of this technique with a simple example from genetics, and an account of our work in language understanding.